\documentclass[10pt,twocolumn,letterpaper]{article}

\usepackage[pagenumbers]{cvpr}

\usepackage{graphicx}
\usepackage{amsmath}
\usepackage{amssymb}
\usepackage{booktabs}
\usepackage{makecell,multirow,diagbox}
\usepackage{subcaption}
\usepackage{epstopdf}
\usepackage[pagebackref,breaklinks,colorlinks]{hyperref}

\usepackage[capitalize]{cleveref}
\crefname{section}{Sec.}{Secs.}
\Crefname{section}{Section}{Sections}
\Crefname{table}{Table}{Tables}
\crefname{table}{Tab.}{Tabs.}

\usepackage{xcolor}

\newcommand\blfootnote[1]{%
\begingroup
\renewcommand\thefootnote{}\footnote{#1}%
\addtocounter{footnote}{-1}%
\endgroup
}

\begin{document}

\title{
SwinTextSpotter: Scene Text Spotting via Better Synergy between Text Detection and Text Recognition
}

\author{
Mingxin Huang\textsuperscript{1}$^\dag$
\quad Yuliang Liu\textsuperscript{2}$^\dag$
\quad Zhenghao Peng\textsuperscript{2}
\quad Chongyu Liu\textsuperscript{1}
\quad Dahua Lin\textsuperscript{2} 
\\
\quad Shenggao Zhu\textsuperscript{3}
\quad Nicholas Yuan\textsuperscript{3}
\quad Kai Ding\textsuperscript{4}
\quad Lianwen Jin\textsuperscript{1,5}$^*$
\\
\textsuperscript{1}{South China University of Technology} 
\quad \textsuperscript{2}{Chinese University of Hong Kong}
\quad \textsuperscript{3}{Huawei Cloud AI}
\\
\quad \textsuperscript{4}{IntSig Information Co., Ltd}
\quad \textsuperscript{5}{Peng Cheng Laboratory}
\\
{\tt\small eelwjin@scut.edu.cn}
}

\maketitle

\begin{abstract}

End-to-end scene text spotting has attracted great attention in recent years due to the success of excavating the intrinsic synergy of the scene text detection and recognition. However, recent state-of-the-art methods usually incorporate detection and recognition simply by sharing the backbone, which does not directly take advantage of the feature interaction between the two tasks.
In this paper, we propose a new end-to-end scene text spotting framework termed SwinTextSpotter.
Using a transformer encoder with dynamic head as the detector, we unify the two tasks with a novel Recognition Conversion mechanism to explicitly guide text localization through recognition loss.
The straightforward design results in a concise framework that requires neither additional rectification module nor character-level annotation for the arbitrarily-shaped text. 
Qualitative and quantitative experiments on multi-oriented datasets RoIC13 and ICDAR 2015, arbitrarily-shaped datasets Total-Text and CTW1500, and multi-lingual datasets ReCTS (Chinese) and VinText (Vietnamese) demonstrate SwinTextSpotter significantly outperforms existing methods. Code is available at \url{https://github.com/mxin262/SwinTextSpotter}.

\end{abstract}
\blfootnote{$^\dag$Equal contribution.}
\blfootnote{$^*$Corresponding author.}

\section{Introduction}
\label{sec:intro}
Scene text spotting, which aims to detect and recognize the entire word or sentence in natural images, has raised a lot of attention due to its wide range of applications in autonomous driving~\cite{9551780}, intelligent navigation~\cite{wang2015bridging,rong2016guided}, and key entities recognition~\cite{zhang2020trie,wang2021towards}, \textit{etc}. Traditional scene text spotting methods treat detection and recognition as two separate tasks and adopt a pipeline that first localizes and crops the text regions on the input images and then predicts the text sequence by feeding the cropped regions into text recognizer \cite{jaderberg2016reading,liao2018textboxes++,jaderberg2014deep,neumann2015real,gomez2017textproposals}. Such a pipeline may have some limitations, such as (1) error accumulation between these two tasks, e.g., imprecise detection result may heavily hinder the performance of text recognition; 
(2) separate optimization of the two tasks might not maximizing the final performance of text spotting; 
(3) intensive memory consumption and low inference efficiency.

\begin{figure}[t!]
    \centering
    \subcaptionbox{Without Recognition Conversion}{
    \includegraphics[width=3.8cm,height=2.5cm]{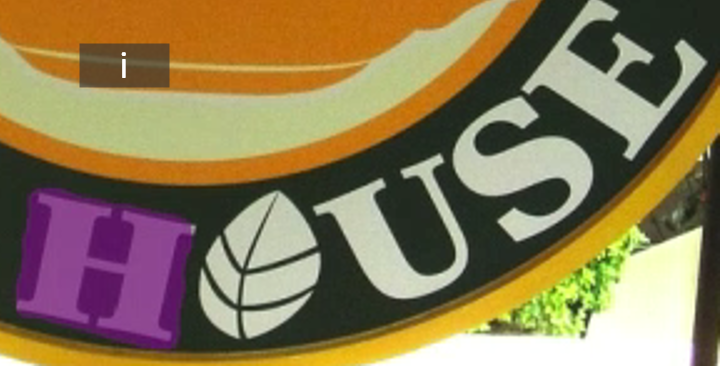}
    \label{fig:Without Recognition Conversion}
    }
    \subcaptionbox{With Recognition Conversion }{\includegraphics[width=3.8cm,height=2.5cm]{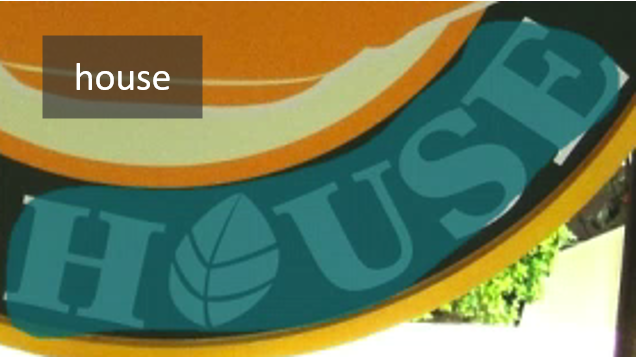}\label{fig:With Recognition Conversion}}
    \subcaptionbox{
    The recognition loss with and without Recognition Conversion.
    }{\includegraphics[width=7.6cm]{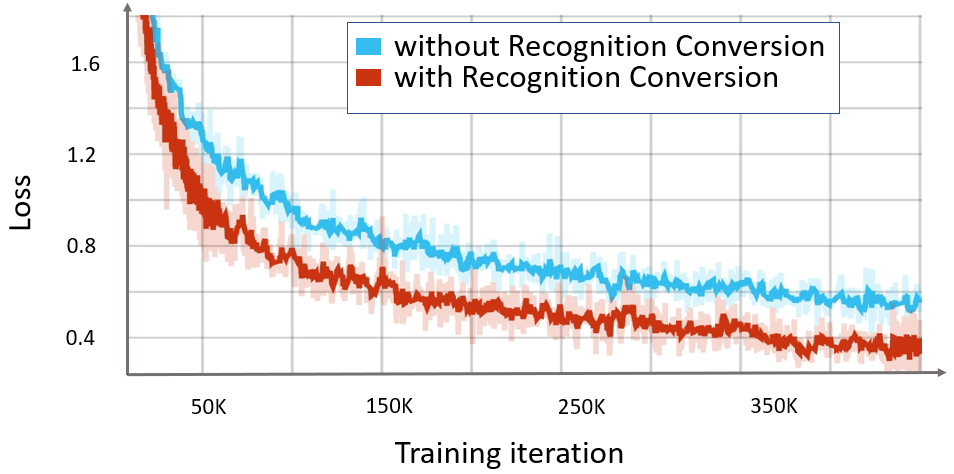}\label{fig:loss}}
        \caption{Effectiveness of Recognition Conversion. The proposed Recognition Conversion explicitly guides the detection, leading to better text spotting performance.
        }
    \label{fig:Compared with Recognition Conversion and without Recognition Conversion}
\end{figure}

Therefore, many methods \cite{li2017towards,he2018end,liu2018fots,lyu2018mask} attempt to solve text spotting in end-to-end systems, \textit{i.e.}, optimizing detection and recognition jointly in unify architectures.
The recognizer can improve the performance of the detector by eliminating the false positive detection results\cite{li2017towards,liao2019mask}. 
In turn, even if the detection is not precise, the recognizer can still correctly predict the text sequence by the large receptive field of the feature map\cite{liu2018fots,liao2019mask}. 
Another advantage is that an end-to-end system is easier to maintain and transfer to new domains compared to a cascaded pipeline where the model is coupled with the data and thus requires substantial engineering efforts\cite{liu2021abcnetv2,wang2021pan++}.

\begin{figure*}[htp]
    \centering
    \includegraphics[width=\textwidth]{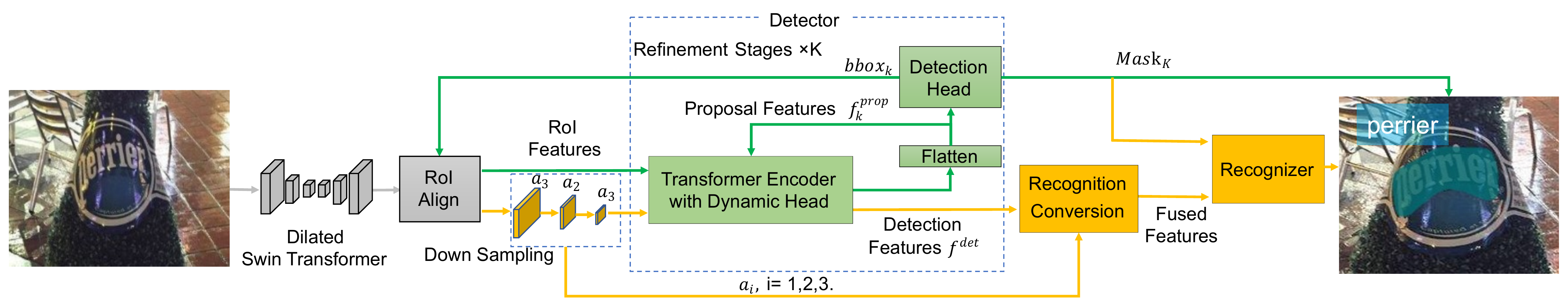}
    \caption{
        The framework of the proposed SwinTextSpotter. The gray arrows denote the feature extraction from images. The green arrows and orange arrows represent the detection stage and the recognition stage, respectively. The outputs of detection head are refined in K stages. The output detection in the $K^{th}$ stage serves as the input to the recognition stage.
    }
    \label{fig:frame}

\end{figure*}

However, there are two limitations in most of the existing end-to-end scene text spotting systems \cite{lyu2018mask,feng2019textdragon,qin2019towards,qiao2020text,wang2020all,liao2020mask,liu2020abcnet}.First, if the detection is simply based on the visual information in the input features, the detector is prone to being distracted by background noise and proposes inconsistent detection, as depicted in Figure \ref{fig:Compared with Recognition Conversion and without Recognition Conversion}(a).
The interaction between texts in the same image is the crucial factor to eliminate the impact of background noise, since different characters of the same word may contain strong similarity, such as the backgrounds and text styles. Using Transformer \cite{vaswani2017attention} can learn rich interactions between text instances.
For example, Yu et al. \cite{yu2020towards} use transformer to make texts interact with each other at semantic level. Fang et al. \cite{fang2021read} and Wang et al. \cite{wang2021two} further adopt transformer to model the visual relationship between texts.
Second, the interactions between detection and recognition is not enough by sharing backbone because neither the detector optimizes the recognition loss nor the recognizer utilizes the detection features.
To jointly improve detection and recognition, 
a character segmentation map is designed by Mask TextSpotter \cite{liao2020mask}, which simultaneously optimizes the detection and recognition results in the same branch; 
ABCNet v2 \cite{liu2021abcnetv2} proposes Adaptive End-to-End Training (AET) strategy using the detection results to extract recognition features instead of only using the ground truths;
ARTS \cite{zhong2021arts} improves the performance of the end-to-end text spotting by back-propagating the loss from the recognition branch to the detection branch using a differentiable Spatial Transform Network (STN)~\cite{2015Spatial}. 
However, these three methods assume the detector proposes text features structurally, \textit{e.g.} in the reading order. The overall performance of the text spotting is thereafter bounded by the detector.

We propose \textit{SwinTextSpotter}, an end-to-end trainable Transformer-based framework, stepping toward better synergy between the text detection and recognition.
To better distinguish the densely scattered text instances in crowded scenes, we use Transformer and a two-level self-attention mechanism in SwinTextSpotter, stimulating the interactions between the text instances.
Addressing the challenge in arbitrarily-shaped scene text spotting, inspired by \cite{sun2021sparse,hu2021istr}, we regard text detection task as a set-prediction problem and thus adopt a query-based text detector.
We further propose \textit{Recognition Conversion (RC)}, which implicitly guides the recognition head through incorporating the detection features.
RC can back-propagate recognition information to the detector and suppress the background noise in the features for recognition, leading to the joint optimization of the detector and recognizer.
Empowered by the proposed RC, SwinTextSpotter has a concise framework without the character-level annotation and rectification module used in previous works to improve the recognizer.
SwinTextSpotter has superior performance in both the detection and the recognition.
As illustrated in Figure \ref{fig:Compared with Recognition Conversion and without Recognition Conversion}(b), the detector of SwinTextSpotter can accurately localize difficult samples.
On the other hand, more accurate detection features can improve the recognizer and result in faster convergence and better performance, as shown in Figure \ref{fig:Compared with Recognition Conversion and without Recognition Conversion}(c).

We conduct extensive experiments on six benchmarks, including multi-oriented dataset RoIC13 \cite{liao2020mask} and ICDAR 2015 \cite{karatzas2015icdar}, arbitrarily-shaped dataset Total-Text \cite{ch2020total} and SCUT-CTW1500 \cite{liu2019curved}, and multilingual dataset ReCTS (Chinese) \cite{zhang2019icdar} and VinText (Vietnamese) \cite{m_Nguyen-etal-CVPR21}. The results demonstrate the superior performance of the SwinTextSpotter: 
(1) SwinTextSpotter achieves 88.0\% F-measure for the detection task on SCUT-CTW1500 and Total-Text, exceeding previous methods by a large margin; 
(2) SwinTextSpotter significantly outperforms ABCNet v2 \cite{liu2021abcnetv2} by 9.8\% in terms of 1-NED for the text spotting task in ReCTs dataset.
Additionally, without using character-level annotation on ReCTs, SwinTextSpotter outperforms previous state-of-the-art methods MaskTextSpotter \cite{liao2019mask} and AE TextSpotter \cite{wang2020ae} that use such annotation; 
(3) SwinTextSpotter shows better robustness for the extremely rotated instances on RoIC13 dataset compared to MaskTextSpotter v3 \cite{liao2019mask}.
The main contributions of this work are summarized as follows.
\begin{itemize}
\item SwinTextSpotter groundbreakingly shows that Transformer and the set-prediction scheme are effective in end-to-end scene text spotting.
\item SwinTextSpotter adopts the \textit{Recognition Conversion} to exploit the synergy of text detection and recognition.
\item SwinTextSpotter is a concise framework that does not require character-level annotation as well as specifically designed rectification module for recognizing arbitrarily-shaped text.
\item SwinTextSpotter achieves state-of-the-art performance on multiple public scene text benchmarks. 
\end{itemize}

\section{Related Work}

\textbf{Separate Scene Text Spotting}.
In past decades, the emergence of deep learning approaches greatly promote the development of scene text spotting. Wang et al. \cite{wang2011end} use a sliding-window-based detector to detect characters and then classify each character. Bissacco et al. \cite{bissacco2013photoocr} combine DNN and HOG features and build a text extraction system by using characters classification. Liao et al. \cite{liao2017textboxes} propose the TextBoxes that incorporates the single-shot detector and a text recognizer \cite{shi2016end} in two-stage manner. However, the aforementioned methods treat detection and recognition as separate tasks without exchange of information between the two tasks.

\textbf{End-to-End Text Spotting}.
Recently, researchers try to combine detection and recognition into one system. Li et al. \cite{li2017towards} unify the detection and recognition into an end-to-end trainable scene text spotting framework.
FOTS \cite{liu2018fots} uses an one stage detector to generate a rotated boxes and adopts RoIRotate to sample the oriented text feature into horizontal grid for connecting the detection and recognition. He et al. \cite{he2018end} propose a similar framework using an attention-based recognizer.

For the task of arbitrarily-shaped scene text spotting, 
Mask TextSpotter series \cite{lyu2018mask,liao2019mask,liao2020mask} solve the problem without explicit rectification by using character segmentation branch to improve the performance of the recognizer.
TextDragon \cite{feng2019textdragon} combines the two tasks by RoISlide, a technique transforms the predicting segments of the text instances into horizontal features. Wang et al. \cite{wang2020all} adopt Thin-Plate-Spline\cite{bookstein1989principal} transformation to rectify the features. ABCNet \cite{liu2020abcnet} and its improved version ABCNet v2~\cite{liu2021abcnetv2} use the BezierAlign to transform the arbitrary-shape texts into regular ones. These methods achieve great progress by using rectification module to unify detection and recognition into end-to-end trainable systems. 
Qin et al.\cite{qin2019towards} propose RoI Masking to extract the feature for arbitrarily-shaped text recognition. Similar to \cite{qin2019towards}, PAN++\cite{wang2021pan++} is based on a faster detector\cite{wang2019efficient}. AE TextSpotter\cite{wang2020ae} uses the results of recognition to guide detection through language model. Though achieve significantly improvement on the performance of text spotting by sharing backbone, the aforementioned methods neither back-propagate recognition loss to the detector nor use detection features in the recognizer. The detector and the recognizer thus are still relatively independent to each other without joint optimization.
Recently, Zhong et al. \cite{zhong2021arts} propose ARTS which passes the gradient of recognition loss to the detector using Spatial Transform Network (STN)\cite{2015Spatial}, demonstrating the power of synergy between the detection and recognition in text spotting.

\section{Methodology}
\label{sec:method}

The overall architecture of SwinTextSpotter is presented in Figure \ref{fig:frame}, which consists of four components: 
(1) a backbone based on Swin-Transformer \cite{liu2021Swin};
(2) a query-based text detector; 
(3) a \textit{Recognition Conversion} module to bridge the text detector and recognizer; and 
(4) an attention-based recognizer. 

As illustrated in the green arrows of Figure~\ref{fig:frame}, in the first stage of detection, we first randomly initialize trainable parameters to be the boxes $bbox_0$ and proposal features $f_0^{prop}$. To make the proposal features contain global information, we use global average pooling to extract the image features and add them into $f_0^{prop}$.
We then extract the RoI features using $bbox_0$. 
The RoI features and $f_0^{prop}$ are fed into the Transformer encoder with dynamic head.
The output of the Transformer encoder is flattened and forms the proposal features $f_1^{prop}$, which will be fed into the detection head to output the detection result. 
The box $bbox_{k-1}$ and proposal feature $f_{k-1}^{prop}$ will serve as the input to later $k^{th}$ stage of detection. 
The proposal feature $f_k^{prop}$ recurrently updates itself by fusing the RoI features with previous $f_{k-1}^{prop}$, which makes proposal features preserve the information from previous stages. 
We repeat such refinement for totally K stages, resembling the iterative structure in the query-based detector~\cite{carion2020end,zhu2020deformable,sun2021sparse,hu2021istr}.
Such design allows more robust detection in sizes and aspect ratios~\cite{sun2021sparse}. 
More details of the detector is explained in Section \ref{sec:A Query Based Detector}.

Since the recognition stage (orange arrows) requires higher rate of resolution than detection, we use the final detection stage output box $bbox_K$ to obtain the RoI features whose resolution is four times as much as that in the detection stage. 
In order to keep the resolution of features consistent with the detector when fused with proposal features, we down-sample the RoI features to get three feature maps of descending sizes, denoting by $\{a_1,a_2,a_3\}$. 
Then we obtain detection features $f^{det}$ by fusing the smallest $a_3$ and the proposal features $f_K^{prop}$.The detection features $f^{det}$ in recognition stage contain all previous detection information. Finally the $\{a_1,a_2,a_3\}$ and the detection features $f^{det}$ are sent into \textit{Recognition Conversion} and recognizer for generating the recognition result.
More details of \textit{Recognition Conversion} and recognizer are explained in Section \ref{sec:Recognition Conversion} and Section \ref{sec:recognizer}, respectively.

\begin{figure}[t]
    \centering
    \includegraphics[width=0.8\linewidth]{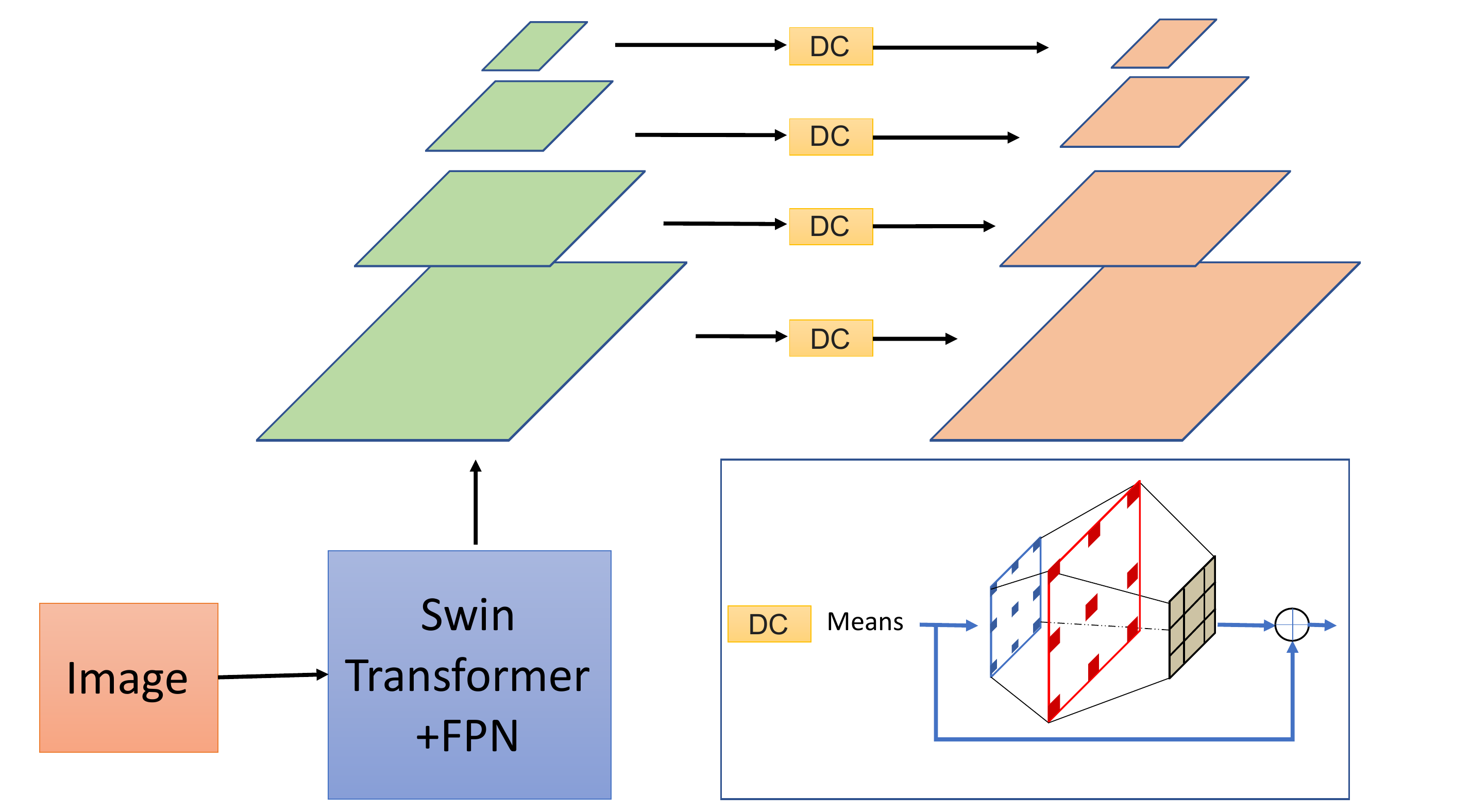}
    \caption{Illustration of the designed Dilated Swin-Transformer. The DC refers to two dilated convolution layers, one vanilla convolution layers and one residual structure.
    }
    \label{fig:Dilated Swin-Transformer}
\end{figure}

\subsection{Dilated Swin-Transformer}

Vanilla convolutions operate locally at fixed size (e.g. $3 \times 3$), which causes low efficacy in connecting remote features. For text spotting, however, modeling the relationships between different texts is critical since scene texts from the same image share strong similarity, such as their backgrounds and text styles. Considering the global modeling capability and computational efficiency, we choose Swin-Transformer \cite{liu2021Swin} with a Feature Pyramid Network (FPN)\cite{lin2017feature} to build our backbone. 
Given the blanks existing between words in a line of text, the receptive field should be large enough to help distinguish whether adjacent texts belong to the same text line. To achieve such receptive field, as illustrated in Figure \ref{fig:Dilated Swin-Transformer}, 
we incorporate two dilated convolution layers\cite{YuKoltun2016}, one vanilla convolution layers and one residual structure into the original Swin-Transformer, which also introduce the properties of CNN to Transformer~\cite{wu2021cvt}.

\subsection{A Query Based Detector}
\label{sec:A Query Based Detector}
We use a query based detector to detect the text. Based on Sparse R-CNN \cite{sun2021sparse}, the query based detector is built on ISTR \cite{hu2021istr} which treats detection as a set-prediction problem. Our detector uses a set of learnable proposal boxes, alternative to replace massive candidates from the RPN~\cite{ren2015faster}, and a set of learnable proposal features, representing high-level semantic vectors of objects. 
The detector is empirically designed to have six query stages.With the Transformer encoder with dynamic head, latter stages can access the information in former stages stored in the proposal features~\cite{jia2016dynamic,tian2020conditional,sun2021sparse}.
Through multiple stages of refinement, the detector can be applied to text at any scale.

\begin{figure}[t!]
    \centering
    \includegraphics[width=\linewidth]{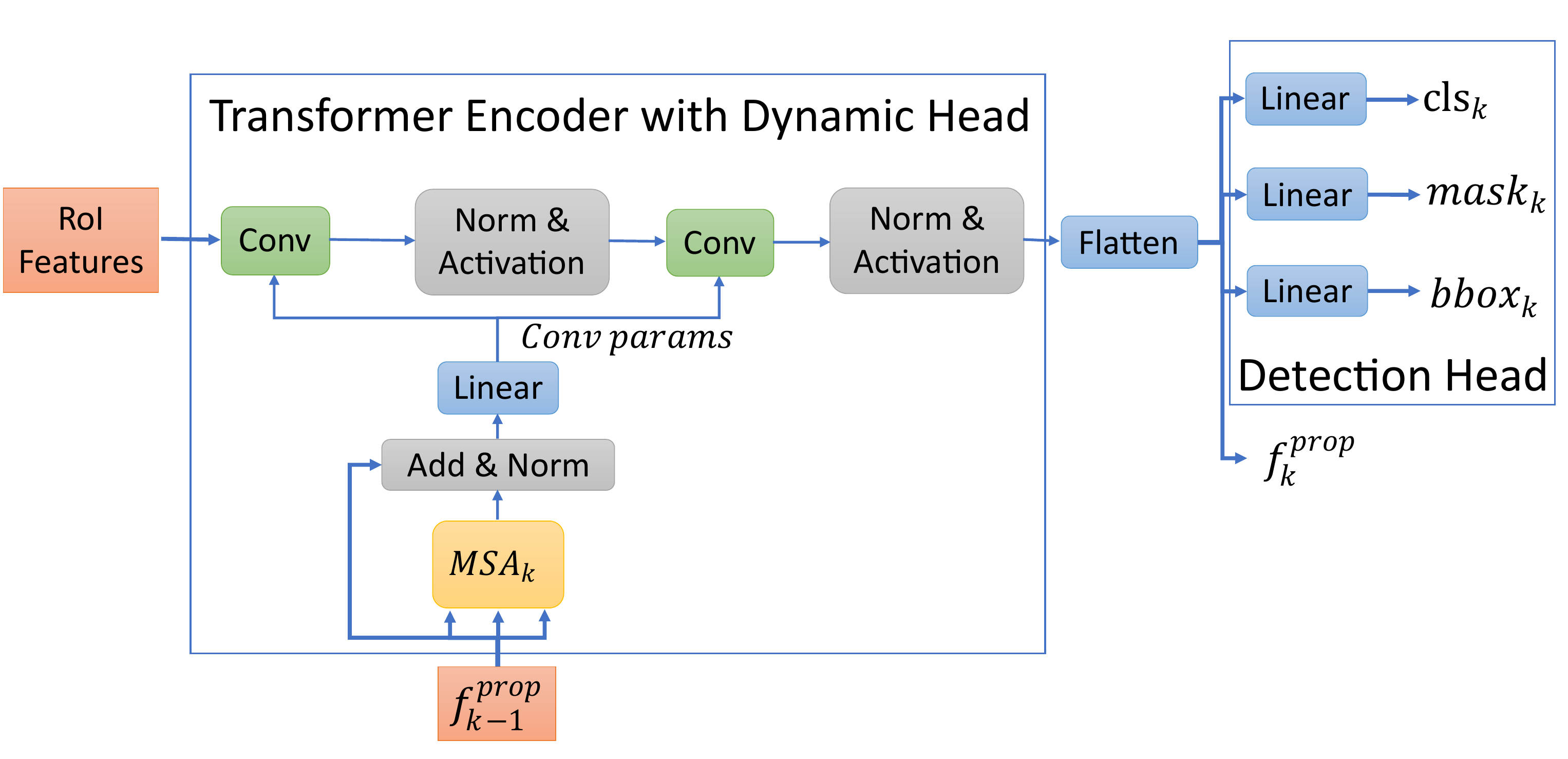}
    \caption{Illustration of $k^{th}$ stage in detection. $f_{k-1}^{prop}$ is the proposal features output by previous stage. 
    $MSA_{k}$ refers to the multi-head-attention in $k^{th}$ stage. $f_{k}^{prop}$ will serve as the input to next stage.
    }
    \label{fig:detector}
\end{figure}

The architecture of the detection head in $k^{th}$ stage is illustrated in Figure \ref{fig:detector}.The proposal features in $k-1$ stage is represented by $f_{k-1}^{prop} \in \mathbb R^{N, d}$.
At the stage $k$, 
the proposal features $f_{k-1}^{prop}$ produced in previous stage is fed into a self-attention module\cite{vaswani2017attention} $MSA_{k}$ to model the relationships and generate two sets of convolutional parameters.
The detection information in previous stages is therefore embedded into the convolutions.The convolutions conditioned on the previous proposal features is used to encode the RoI features.
The RoI features are extracted by $bbox_{k-1}$, the detection result in previous stage, using RoIAlign\cite{he2017mask}.The output features of the convolutions is fed into a linear projection layer to produce the $f_{k}^{prop}$ for next stage. 
The $f_{k}^{prop}$ is subsequently fed the into prediction head to generate $bbox_{k}$ and $mask_{k}$.
To reduce computation, the 2D mask is transformed into 1D mask vector by the Principal Component Analysis\cite{wold1987principal} so the $mask_{k}$ is an one-dimensional vector.

When $k=1$, the $bbox_{0}$ and $f_{0}^{prop}$ are randomly initialized parameters, which is the input of the first stage. 
During training, these parameters are updated via back propagation and learn the inductive bias of the high-level semantic features of text.

We view the text detection task as a set-prediction problem.Formally, we use the bipartite match to match the predictions and ground truths~\cite{carion2020end,stewart2016end,sun2021sparse,hu2021istr}. The matching cost becomes:
\begin{equation}
L_{match} = \lambda_{cls} \cdot L_{cls} + \lambda_{L1} \cdot L_{L1} + \lambda_{giou} \cdot L_{giou} + \lambda_{mask} \cdot L_{mask},\label{XX}
\end{equation}
where $\lambda$ is the hyper-parameter used to balance the loss. $L_{cls}$ is the focal loss\cite{lin2017focal}. 
The losses for regressing the bounding boxes are L1 loss $L_{L1}$ and generalized IoU loss $L_{giou}$ \cite{rezatofighi2019generalized}. 
We compute the mask loss $L_{mask}$ following \cite{hu2021istr}, which calculates the cosine similarity between the prediction mask and ground truth. The detection loss is similar to the matching cost but we use the L2 loss and dice loss \cite{milletari2016v} to replace the cosine similarity as in \cite{hu2021istr}.

\subsection{Recognition Conversion}
\label{sec:Recognition Conversion}

\begin{figure}[t!]
    \centering
    \includegraphics[width=\linewidth]{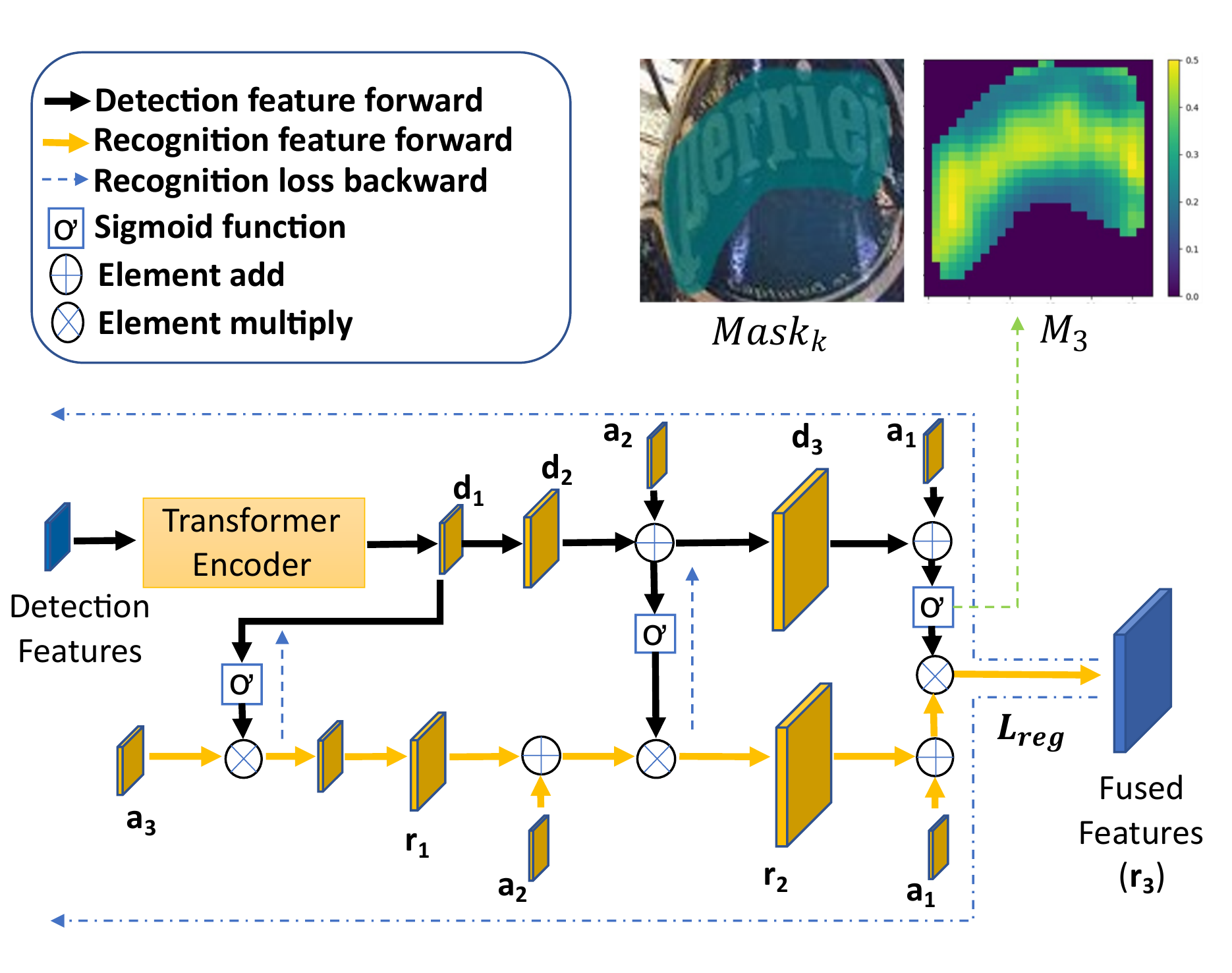}
    \caption{
        Detailed structure of \textit{Recognition Conversion}.
    }
    \label{fig:recognition conversion}
\end{figure}

To better coordinate the detection and recognition, we propose \textit{Recognition Conversion (RC)} to spatially inject the features from detection head into the recognition stage, as illustrated in Figure \ref{fig:recognition conversion}. 
The RC consists of the Transformer encoder \cite{vaswani2017attention} and four up-sampling structures. The input of RC are the detection features $f^{det}$ and three down-sampling features $\{a_1,a_2,a_3\}$.

The detection features are sent to the Transformer encoder $TrE()$, making the information of previous detection stages further fused with $a_3$. 
Then through a stack of up-sampling operation $E_u()$ and Sigmoid function $\phi{()}$, three masks $\{M_1,M_2,M_3\}$ for text regions are generated:
\begin{eqnarray}
d_1 = & TrE(f^{det}), \\
d_2 = & (E_u(d_1) + a_2), \\
d_3 = & (E_u(d_2) + a_1), \\
M_i = & \phi{(d_i)}, i=1,2,3.
\end{eqnarray}

With the masks $\{M_1,M_2,M_3\}$ and the input features $\{a_1,a_2,a_3\}$, we further integrate these features effectively under the following pipeline:
\begin{eqnarray}
r_1 = & M_1 \cdot a_3, \\
r_2 = & M_2 \cdot (E_u(r_1)+a_2), \\
r_3 = & M_3 \cdot (E_u(r_2)+a_1),
\end{eqnarray}

where $\{r_1,r_2,r_3\}$ denote the recognition features. The $r_3$ is the fused features in Figure \ref{fig:recognition conversion}, which is finally sent to the recognizer at the highest resolution. 
As shown in the blue dashed lines in Figure~\ref{fig:recognition conversion},
the gradient of the recognition loss $L_{reg}$ can be back-propagated to the detection features, enabling RC to implicitly improve the detection head through the recognition supervision.

Generally, to suppress the background, the fused features will be multiplied by a $mask_{K}$ predicted by detection head (with the supervision of $L_{mask}$).
However, the background noise still remains in the feature maps as the detection box is not tight enough.
Such issue can be alleviated by the proposed RC since RC uses the detection features to generate tight masks to suppress the background noise, which is supervised by the recognition loss apart from the detection loss. 
As shown in the upper right corner of Figure \ref{fig:recognition conversion}, $M_3$ suppresses more background noise than $mask_{K}$, where $M_3$ has higher activation in texts region and lower in the background.
Therefore the masks $\{M_1,M_2,M_3\}$ produced by RC, which will be applied to the recognition features $\{r_1,r_2,r_3\}$, makes recognizer easier to concentrate on the text regions. 

With RC, the gradient of recognition loss not only flows back to the backbone network, but also to the proposal features. 
Optimized by both detection supervision and recognition supervision, the proposal features can better encode the high-level semantic information of the texts.
Therefore, the proposed RC can incentivize the coordination between detection and recognition. 

\subsection{Recognizer}
\label{sec:recognizer}

After applying RC on the feature map, background noise is effectively suppressed and thus the text regions can be bounded more precisely. 
This enables us to merely use a sequential recognition network to obtain promising recognition results without rectification modules such as TPS \cite{bookstein1989principal}, RoISlide \cite{feng2019textdragon}, Bezier-Align \cite{liu2020abcnet} or character-level segmentation branch used in MaskTextSpotter\cite{liao2019mask}. 
To enhance the fine-grained feature extraction and sequence modeling, we adopt a bi-level self-attention mechanism, inspired by \cite{yang2021focal}, as the recognition encoder. The two-level self-attention mechanism (TLSAM) contains both fine-grained and coarse-grained self-attention mechanisms for local neighborhood regions and global regions, respectively. Therefore, it can effectively extract fine-grained features while maintaining global modeling capability.
As for the decoder, we simply follow MaskTextSpotter by using the Spatial Attention Module (SAM) \cite{liao2020mask}. The recognition loss is as follow:
\begin{equation}
L_{reg} = -\frac{1}{T} \sum_{k=1}^{T}log\ p(y_i),
\end{equation}
wherein $T$ is the max length of the sequence and $p(y_i)$ is the probability of sequence.

\section{Experiments}
We conduct experiments on various scene text benchmarks, including multi-oriented scene text benchmarks RoIC13\cite{liao2020mask} and ICDAR 2015\cite{karatzas2015icdar}, multilingual datasets ReCTS\cite{zhang2019icdar} and Vintext\cite{m_Nguyen-etal-CVPR21}, and two arbitrarily-shaped scene text benchmarks Total-Text\cite{ch2020total} and SCUT-CTW 1500\cite{liu2019curved}. The ablation studies are conducted on Total-Text to verify each component of our proposed method. Unless specified, all values in the table are in percentage.

\subsection{Implementation Details}
\label{Implementation Details}

We follow the training strategy in \cite{qiao2021mango}. First, the model is pretrained on the Curved SynthText \cite{liu2020abcnet}, ICDAR-MLT \cite{nayef2017icdar2017}, and the corresponding dataset for 450K iterations. The initialized learning rate is $2.5 \times 10 ^ {-5}$, which reduces to $2.5 \times 10 ^ {-6}$ at $380K ^ {th}$ iteration and $2.5 \times 10 ^ {-7}$ at $420K ^ {th}$ iteration. Then we jointly train the pretrained model for 80K iterations on the Total-Text, ICDAR 2013, and ICDAR-MLT, which decays to a tenth at 60K. Finally, we fine-tune the jointly trained model on the corresponding datasets. We also follow the training strategies in \cite{liu2021abcnetv2} and \cite{m_Nguyen-etal-CVPR21} to train the model on Chinese and Vietnamese.

We extract 4 feature maps with 1/4, 1/8, 1/16, 1/32 resolution of the input image for text detection and recognition. 

We train our model with image batch size of 8. 
The following data augmentation strategies are used:  
(1) random scaling; (2) random rotation; and (3) random crop. 
Other strategies such as random brightness, contrast, and saturation are also applied during training.

\begin{table}[t]
\begin{tabular}{c|ccc}
\hline
\multirow{2}{*}{Method} & \multicolumn{3}{c}{ICDAR 2015 End-to-End}                    \\ \cline{2-4} 
                        & \multicolumn{1}{c|}{S}    & \multicolumn{1}{c|}{W}    & G    \\ \hline
FOTS\cite{liu2018fots}                    & \multicolumn{1}{c|}{81.1} & \multicolumn{1}{c|}{75.9} & 60.8 \\
Mask TextSpotter\cite{liao2019mask}        & \multicolumn{1}{c|}{83.0} & \multicolumn{1}{c|}{77.7} & 73.5 \\
CharNet\cite{xing2019convolutional}                 & \multicolumn{1}{c|}{83.1} & \multicolumn{1}{c|}{\textbf{79.2}} & 69.1 \\
TextDragon\cite{feng2019textdragon}              & \multicolumn{1}{c|}{82.5} & \multicolumn{1}{c|}{78.3} & 65.2 \\
Mask TextSpotter v3\cite{liao2020mask}     & \multicolumn{1}{c|}{83.3} & \multicolumn{1}{c|}{78.1} & \textbf{74.2} \\
MANGO\cite{qiao2021mango}                   & \multicolumn{1}{c|}{81.8} & \multicolumn{1}{c|}{78.9} & 67.3 \\
PAN++\cite{wang2021pan++}                   & \multicolumn{1}{c|}{82.7} & \multicolumn{1}{c|}{78.2} & 69.2 \\
ABCNet v2\cite{liu2021abcnetv2}                & \multicolumn{1}{c|}{82.7} & \multicolumn{1}{c|}{78.5} & 73.0 \\ \hline
SwinTextSpotter         & \multicolumn{1}{c|}{\textbf{83.9}} & \multicolumn{1}{c|}{77.3} & 70.5 \\ \hline
\end{tabular}
\caption{End-to-end recognition result on ICDAR 2015. “S”, “W”, and “G” represent recognition with “Strong”, “Weak”, and “Generic” lexicon, respectively.}
\label{ICDAR 2015 End-to-End recognition result}
\end{table}

\begin{table}[t]
\centering
\begin{tabular}{c|c|c|c|c}
\hline
\multirow{2}{*}{Method} & \multicolumn{3}{c|}{Detection} & \multirow{2}{*}{1-NED} \\ \cline{2-4}
                        & R        & P        & H        &                        \\ \hline
FOTS\cite{liu2018fots}                    & 82.5     & 78.3     & 80.31    & 50.8                   \\ \hline
MaskTextSpotter\cite{liao2019mask}         & 88.8     & 89.3     & 89.0       & 67.8                   \\ \hline
AE TextSpotter\cite{wang2020ae}          & \textbf{91.0}     & 92.6     & \textbf{91.8}     & 71.8                   \\ \hline
ABCNet v2\cite{liu2021abcnetv2}               & 87.5     & 93.6     & 90.4     & 62.7                   \\ \hline
SwinTextSpotter                    & 87.1     & \textbf{94.1}     & 90.4     & \textbf{72.5}                   \\ \hline
\end{tabular}
\caption{End-to-end text spotting result and detection result on ReCTS.}
\label{ReCTS Rsult}
\end{table}

\begin{table*}[t!]
\centering
\setlength{\tabcolsep}{7mm}{
\begin{tabular}{c|c|c|c|c|c|c}
\hline
\multirow{2}*{Method} & \multicolumn{3}{c|}{Rotation Angle $45^{\circ}$}                                            & \multicolumn{3}{c}{Rotation Angle $60^{\circ}$}  \\ \cline{2-7} 
                                                       & R                         & P                         & H                         & R           & P           & H          \\ \hline
CharNet\cite{xing2019convolutional}                                                & 35.5                      & 34.2                      & 33.9                      & 8.4         & 10.3        & 9.3        \\ 
Mask TextSpotter\cite{liao2019mask}                                       & \multicolumn{1}{c|}{45.8} & \multicolumn{1}{c|}{66.4} & \multicolumn{1}{c|}{54.2} & 48.3        & 68.2        & 56.6       \\ 
Mask TextSpotter v3\cite{liao2020mask}                                    & 66.8                      & \textbf{88.5}                      & 76.1                      & 67.6        & \textbf{88.5}        & 76.6       \\ 
SwinTextSpotter                                                   & \textbf{72.5}                      & 83.4                      & \textbf{77.6}                      & \textbf{72.1}        & 84.6        & \textbf{77.9}       \\ \hline
\end{tabular}}
\caption{End-to-end recognition result on RoIC13. P, R, H represent precision,
recall and Hmean, respectively.}
\label{Rotate-IC13-end-to-end}
\end{table*}

\subsection{Datasets}
We use the following datasets:
\textbf{Curved SynthText} \cite{liu2020abcnet}  is a synthesized dataset for arbitrarily-shaped scene text. It contains 94,723 images with multi-oriented text and 54,327 images with curved text.
\textbf{ICDAR 2013} \cite{karatzas2013icdar} is a scene text dataset proposed in 2013. It contains 229 training images and 233 test images.
\textbf{ICDAR 2015 \cite{karatzas2015icdar}} is built in 2015. It contains 1,000 training images and 500 test images.
\textbf{ICDAR 2017} \cite{nayef2017icdar2017} is a multi-lingual text dataset. It contains 7,200 training images, 1,800 validation images.
We only select the images containing English texts for training.
\textbf{ICDAR19 ArT}\cite{chng2019icdar2019} is a dataset for arbitrarily shaped text. It contains 5,603 training images.
\textbf{ICDAR19 LSVT}\cite{sun2019icdar} is a large number of Chinese datasets which contains 30,000 for training images.
\textbf{Total-Text} \cite{ch2020total} is the benchmark for arbitrarily-shaped scene text. It consists of 1,255 training images and 300 testing images.
The word-level polygon boxes are provided as the annotations.
\textbf{SCUT-CTW1500} \cite{liu2019curved} is a text-line level arbitrarily-shaped scene text dataset. It consists of 1,000 training images and 500 testing images. Compared to Total-Text, this dataset contains denser and longer text. 
\textbf{ReCTS} \cite{zhang2019icdar} consists of 20,000 training images and 5,000 testing images. It also provides character-level bounding boxes, which are not used in our method.
\textbf{VinText} \cite{m_Nguyen-etal-CVPR21} is a recently proposed Vietnamese text dataset. It consists of 1,200 training images and 500 testing images.

\begin{table}[t]
\centering
\begin{tabular}{c|c}
\hline
Method & H-mean                       \\ \hline 
ABCNet\cite{liu2020abcnet}                  & 54.2                     \\ \hline
ABCNet+D\cite{m_Nguyen-etal-CVPR21}     & 57.4                     \\ \hline
Mask Textspotter v3\cite{m_Nguyen-etal-CVPR21}     & 53.4                     \\ \hline
Mask Textspotter v3+D\cite{m_Nguyen-etal-CVPR21}     & 68.5                     \\ \hline
SwinTextSpotter                    & \textbf{71.1}                        \\ \hline
\end{tabular}
\caption{End-to-end text spotting result on VinText. ABCNet+D means adding the methods proposed in \cite{m_Nguyen-etal-CVPR21} to ABCNet. The same to Mask Textspotter v3+D.}
\label{VinText}
\end{table}

\begin{table}[h]
\centering
\begin{tabular}{c|c|c|c}
\hline
\multirow{2}*{Method} & Detection & \multicolumn{2}{c}{End-to-End}                      \\ \cline{2-4}
           & H-mean             & None                      & \multicolumn{1}{c}{Full} \\ \hline
CharNet\cite{xing2019convolutional}    & 85.6           & \multicolumn{1}{c|}{66.6} & \multicolumn{1}{c}{-}   \\ 
ABCNet\cite{liu2020abcnet}   & -               & 64.2                      & \multicolumn{1}{c}{75.7}  \\ 
PGNet\cite{wang2021pgnet}    & 86.1           & \multicolumn{1}{c|}{63.1} & \multicolumn{1}{c}{-}   \\ 
Mask TextSpotter\cite{liao2019mask}  & 85.2      & 65.3                      & 77.4                    \\
Qin et al. \cite{qin2019towards}        & -       &67.8                       & -  \\
Mask TextSpotter v3\cite{liao2020mask}      & -   & 71.2                      & 78.4                  \\ 
MANGO\cite{qiao2021mango}      & -        & 72.9                      & 83.6                  \\ 
ABCNet v2\cite{liu2021abcnetv2}     &87.0          & 70.4                      & 78.1                 \\ 
PAN++\cite{wang2021pan++}      & -       & 68.6                      & 78.6                  \\ \hline
SwinTextSpotter-Res      & 87.2             & 72.4                      & \multicolumn{1}{c}{83.0}   \\
SwinTextSpotter       & \textbf{88.0}          & \textbf{74.3}                      & \multicolumn{1}{c}{\textbf{84.1}}   \\ \hline
\end{tabular}
\caption{End-to-end text spotting result and detection result on Total-Text. SwinTextSpotter-Res means using the ResNet50 with FPN as backbone.  ``None"
represents lexicon-free. ``Full" represents that we use all the words appeared in the test set. }
\label{Total-Text e2e}
\end{table}

\subsection{Comparisons with State-of-the-Art methods}
Except in special cases, all values in the table are in percentage.

\begin{figure*}[h]
    \centering
    \subcaptionbox{Total-Text}{\includegraphics[width=5.6cm,height=2.5cm]{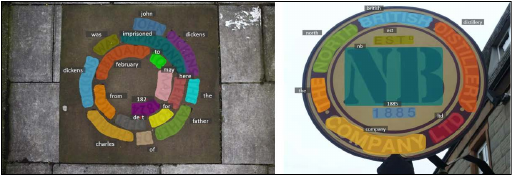}\label{fig:vis_tt}}
    \subcaptionbox{CTW1500}{\includegraphics[width=5.6cm,height=2.5cm]{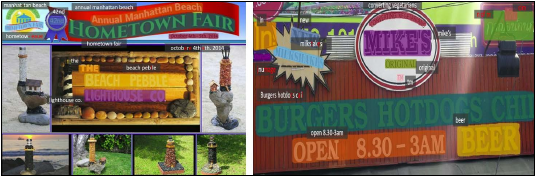}\label{fig:vis_ctw1500}}
    \subcaptionbox{VinText}{\includegraphics[width=5.6cm,height=2.5cm]{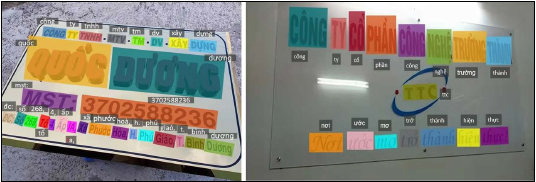}\label{fig:vis_vintext}}
    \subcaptionbox{ReCTS}{\includegraphics[width=5.6cm,height=2.5cm]{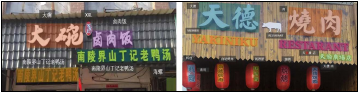}\label{fig:vis_rects}}
    \subcaptionbox{ICDAR2015}{\includegraphics[width=5.6cm,height=2.5cm]{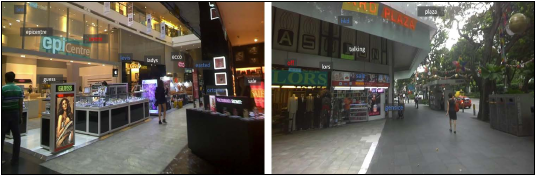}\label{fig:vis_ic15}}
    \subcaptionbox{RoIC13}{\includegraphics[width=5.6cm,height=2.5cm]{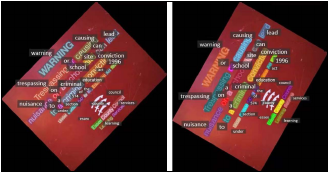}\label{fig:vis_roic13}}
    \caption{Visualization results of our method. White text represents the correct results; Red text represents the wrong results; Blue text represents that the GT of the text instance is marked as ``do not care". Best view in screen.}
    \label{fig:vis_results}
\end{figure*}

\textbf{Multi-oriented and Multilingual datasets}. We first conduct experiments on ICDAR 2015, showing the superiority of SwinTextSpotter on oriented scene text. Table \ref{ICDAR 2015 End-to-End recognition result} shows that SwinTextSpotter achieves the best strong lexicon results on ICDAR 2015, without using the character-level annotations which were used by ABCNet v2 and MaskTextSpotter v3. We also conduct experiments on RoIC13 dataset proposed in \cite{liao2020mask} to verify the rotation robustness of SwinTextSpotter. The end-to-end recognition results are shown in Table \ref{Rotate-IC13-end-to-end}. Both in Rotation Angle $45^{\circ}$ and in Rotation Angle $60^{\circ}$ datasets, SwinTextSpotter can achieve the state-of-the-art in terms of the H-mean metric. Our method significantly outperforms the Mask TextSpotter v3 by 1.5$\%$ in terms of  
H-mean on Rotation Angle $45^{\circ}$ and 1.3$\%$  on Rotation Angle $60^{\circ}$. In addition to English, we also conduct experiment on Chinese dataset ReCTS and Vietnamese dataset VinText to verify the generality of SwinTextSpotter. As shown in Table \ref{ReCTS Rsult}, for ReCTS, our method surpasses ABCNet v2, which only works on word-level annotation, by 9.8$\%$ in 1-NED.
SwinTextSpotter has 0.7$\%$ higher 1-NED than AE-TextSpotter, the SOTA method requiring additional character-level annotation. 
For VinText, the end-to-end results are presented in Table \ref{VinText}, ``D" means using dictionary for the training of recognizer. SwinTextSpotter can also outperform previous methods on VinText, showing the generalization of our method.
It is worth noting that for the above tasks, we do not use dictionary for the training of recognizer as ABCNet+D and Mask TextSpotter v3+D.
The qualitative results of these three benchmarks are shown in Figure \ref{fig:vis_results} (c)(d)(e)(f). 

\begin{table}[t]
\centering
\begin{tabular}{c|c|c|c|c}
\hline
\multirow{2}*{Method} & Detection & \multicolumn{2}{c|}{End-to-End} & \multirow{2}*{1-NED}                      \\ \cline{2-4} 
            & H-mean           & None                      & \multicolumn{1}{c|}{Full} & \\ \hline
TextDragon\cite{feng2019textdragon}       & 83.6           & 39.7                      & 72.4              & -      \\ 
ABCNet\cite{liu2020abcnet}     & 81.4            & 45.2                      & \multicolumn{1}{c|}{74.1} & - \\
MANGO\cite{qiao2021mango}      & -            & \textbf{58.9}                      & \textbf{78.7}                   & - \\ 
ABCNet v2\cite{liu2021abcnetv2}   & 84.7            & 57.5                      & 77.2                  & 46.9  \\ \hline
SwinTextSpotter         & \textbf{88.0}           & 51.8                      & \multicolumn{1}{c|}{77.0}   & 45.7\\ \hline
\end{tabular}
\caption{End-to-end text spotting result and detection result on SCUT-CTW1500.  ``None"
represents lexicon-free. ``Full" represents that we use all the words appeared in the test set.}
\label{CTW1500 e2e}
\end{table}

\begin{table*}[t!]
\centering
\setlength{\tabcolsep}{0.5mm}{
\begin{tabular}{c|cccc|cc}
\hline
Method & \multicolumn{1}{l|}{\multirow{2}{*}{Recognition Conversion}} & \multicolumn{1}{l|}{\multirow{2}{*}{Swin-Transformer}} & \multicolumn{1}{l|}{\multirow{2}{*}{TLSAM}} & \multicolumn{1}{l|}{\multirow{2}{*}{Dilated convolution}} & \multicolumn{2}{c}{Total-Text}                                 \\ \cline{6-7} 
                       & \multicolumn{1}{l|}{}                                        & \multicolumn{1}{l|}{}                                  & \multicolumn{1}{l|}{}                                   & \multicolumn{1}{l|}{}                                     & \multicolumn{1}{l|}{Det-Hmean} & \multicolumn{1}{l}{E2E-Hmean} \\ \hline
$Baseline$    &  &  & & & \multicolumn{1}{c|}{78.9}      & 55.7                          \\ \cline{1-1} \hline
$Baseline+$              & \checkmark                                                            &                                                       &                                                        &                                                          & \multicolumn{1}{c|}{81.9}      & 62.6                          \\ \cline{1-1}
$Baseline+$              & \checkmark                                                            & \checkmark                                                      &                                                        &                                                          & \multicolumn{1}{c|}{81.7}      & 65.5                          \\ \cline{1-1}
$Baseline+$              & \checkmark                                                            &                                                       & \checkmark                                                       &                                                          & \multicolumn{1}{c|}{81.7}      & 62.5                          \\ \cline{1-1}
$Baseline+$             & \checkmark                                                            & \checkmark                                                      &                                                        & \checkmark                                                         & \multicolumn{1}{c|}{82.4}      & 66.0                          \\ \cline{1-1} \hline
SwinTextSpotter              & \checkmark                                                            & \checkmark                                                      & \checkmark                                                       & \checkmark                                                         & \multicolumn{1}{c|}{\textbf{83.2}}      & \textbf{66.9}                          \\ \hline

\end{tabular}}
\caption{Ablation studies on Total-Text without finetuning. ResNet-50 is used as the baseline backbone. TLSAM stands for the two-level self-attention mechanism. }
\label{Ablation}
\end{table*}

\textbf{Irregular text}. We conduct experiments on two arbitrarily-shaped scene text datasets (Total-Text and SCUT-CTW1500) for both detection and end-to-end scene text spotting tasks. In text detection task, the results in Table \ref{Total-Text e2e} and \ref{CTW1500 e2e} demonstrate that SwinTextSpotter can achieve 88$\%$ H-mean on both datasets, which outperforms previous state-of-the-art methods 1.0$\%$ and 3.3$\%$ for Total-Text and SCUT-CTW1500, respectively.
As for end-to-end scene text spotting task, according to Table \ref{Total-Text e2e}, SwinTextSpotter significantly outperforms previous methods on Total-Text with 74.3$\%$ F-measure, 3.9$\%$ higher than ABCNet v2 and 1.4$\%$ higher than MANGO. Besides, for fair comparison to previous methods, we replace our Dilated Swin-Transformer backbone with ResNet-50 and the performance is still comparable to the best result (72.4$\%$ in our method and 72.9$\%$ in MANGO). On SCUT-CTW1500, however, as presented in Table \ref{CTW1500 e2e}, though our method can achieve the best performance on text detection, the end-to-end text spotting result still contain a gap. We discuss and analyze such phenomenon in Section \ref{subsec:discuss}. Some qualitative results are shown in Figure \ref{fig:vis_results}(a)(b).

\subsection{Ablation Studies}
To evaluate the effectiveness of the proposed components, we conduct ablation studies on Total-Text. ResNet-50 is used as the baseline backbone. 
In ablation studies, we only train different variants of SwinTextSpotter on the Curved SynthText, ICDAR-MLT and the corresponding dataset as the first stage described in Section \ref{Implementation Details}.

\textbf{Recognition Conversion}. As shown in Table \ref{Ablation}, with RC, the results can be improved by 3.0$\%$ and 6.9$\%$ for detection and end-to-end scene text spotting, respectively. RC greatly improve the performance of the detector and recognizer. This is mainly because the RC can generate more discriminative features for text regions to boost the performance of text recognition so as to benefit the text detector.

\textbf{Dilated Swin-Transformer}. We also compare the performance of different backbones. The model with Swin-Transformer can achieve 2.9$\%$ improvement on end-to-end results over the ResNet-50, but there is no improvement for detection. Incorporating dilated convolution into Swin-Transformer can further boost detection by 0.7$\%$ and 0.5$\%$ in end-to-end results.

\textbf{Two-level self-attention mechanisms}. 
In Table \ref{Ablation}, we further conduct experiments to explore the influence of fine-grained features. 
Dilated Swin-Transformer performs poorly in capturing fine-grained features, while two-level self-attention mechanisms can effectively make up for this shortcoming. The enhancement of fine-grained features from two-level self-attention mechanism can result in 0.8$\%$ and 0.9$\%$ improvement on text detection and end-to-end text spotting results, respectively.

\subsection{Limitation and Discussion}
\label{subsec:discuss}
\textbf{Long Arbitrarily-Shaped Text}. We know that the long arbitrarily-shaped text requires a high resolution feature map to be recognized. When the feature map becomes larger, attention map in the recognizer will also expand. 
Large attention map may result in mismatch of the recognizer, which leads to the low end-to-end text spotting performance on SCUT-CTW1500. The amount of long arbitrarily-shaped data is limited. Our recognition decoder needs more training data than the 1D-Attention\cite{bahdanau2014neural} and CTC\cite{graves2006connectionist} and so it is not well trained yet.
However, the 1-NED result and ``Full" result shown in Table \ref{CTW1500 e2e} narrow the gaps between our method and ABCNet v2, which suggests that the errors mainly occur in individual characters.

\section{Conclusion}
In this paper, we propose SwinTextSpotter. To the best of our knowledge, SwinTextSpotter is the first successful attempt using Transformer-based method and set-prediction scheme for end-to-end scene text spotting. With the core idea to make text recognition as a part of detection, the proposed model tightly couples the detection and recognition instead of only sharing information in the backbone.
The proposed \textit{Recognition Conversion} enables the suppression of background noise in the recognition features by making the detection results differentiable with respect to the recognition loss.
Such design greatly simplifies the text spotter framework by removing the rectification module and enables the joint optimization of both the detection and the recognition module toward better spotting performance without character-level annotation.
Extensive experiments on public benchmarks demonstrate that SwinTextSpotter can achieve superior performance in end-to-end scene text spotting on arbitrarily-shaped text and multi-lingual text.

\noindent \textbf{Acknowledgement} This research is supported in part by NSFC (Grant No.: 61936003) and GD-NSF (No.2017A030312006, No.2021A1515011870).

\appendix
{\centering\section*{Appendix}}
\section{Qualitative Comparisons}
\begin{figure}[h]
    \centering
    \subcaptionbox{MaskTextSpotter}{\includegraphics[width=2.6cm,height=3.0cm]{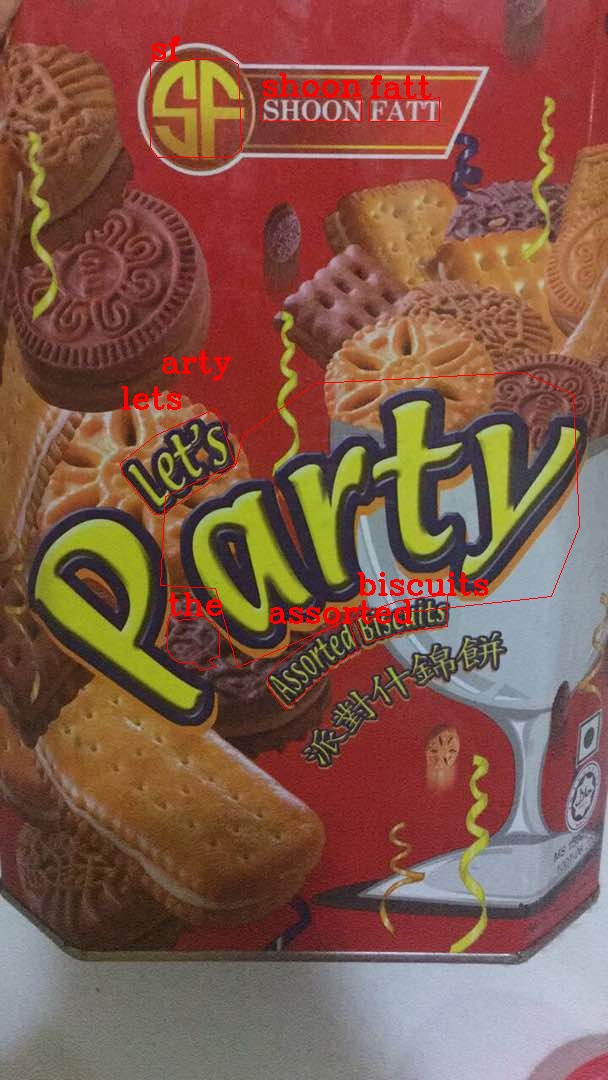}\label{fig:mts}}
    \subcaptionbox{MaskTextSpotterV3}{\includegraphics[width=2.6cm,height=3.0cm]{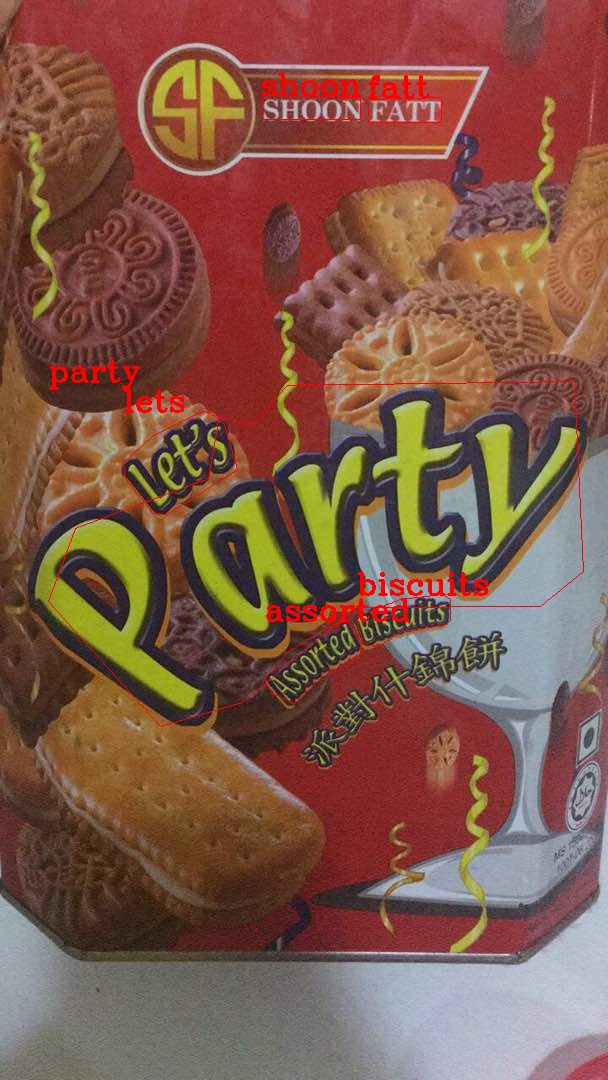}\label{fig:mtsv3}}
    \subcaptionbox{ABCNet}{\includegraphics[width=2.6cm,height=3.0cm]{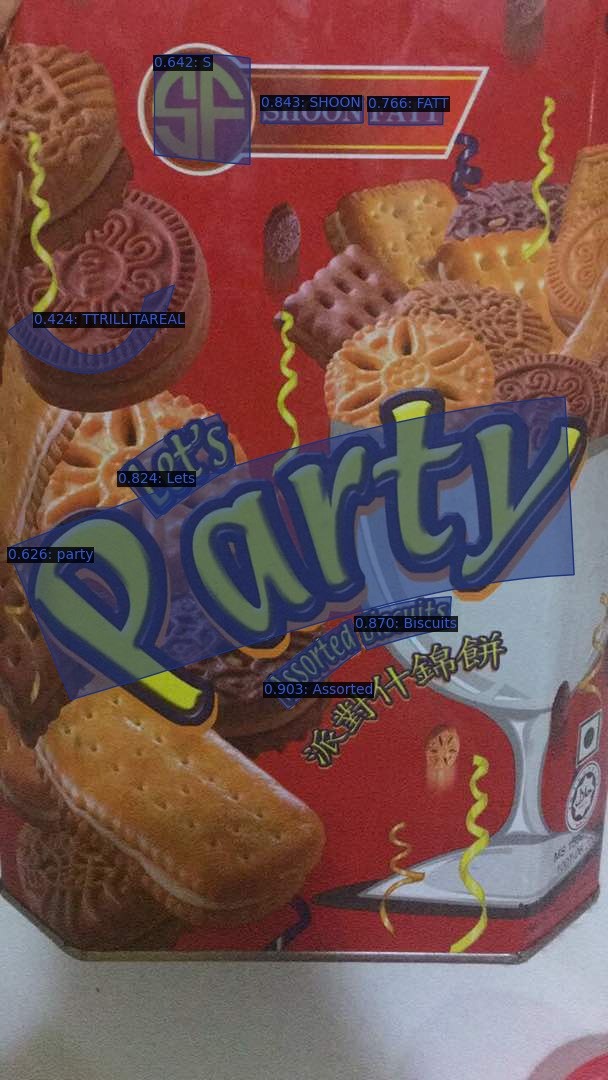}\label{fig:abc}}
    \subcaptionbox{ABCNetv2}{\includegraphics[width=2.6cm,height=3.0cm]{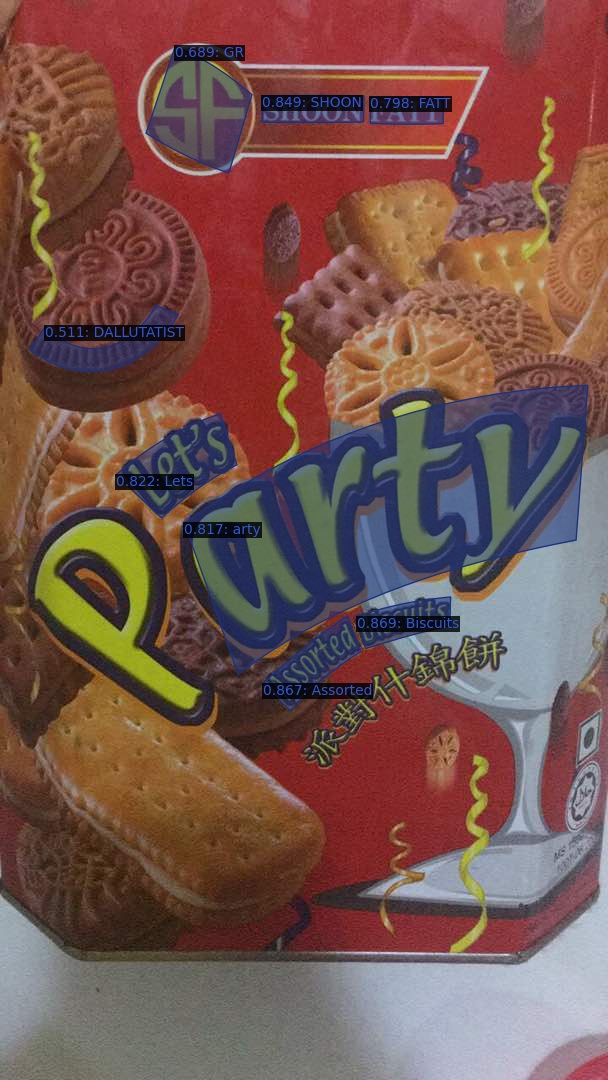}\label{fig:abcv2}}
    \subcaptionbox{MANGO}{\includegraphics[width=2.6cm,height=3.0cm]{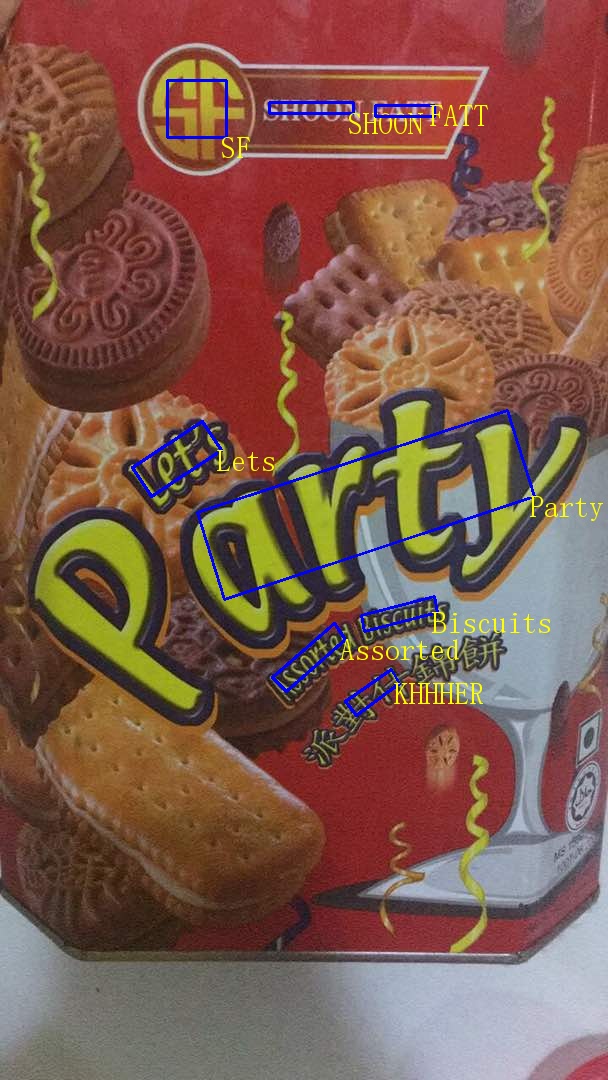}\label{fig:mango}}
    \subcaptionbox{SwinTextSpotter}{\includegraphics[width=2.6cm,height=3.0cm]{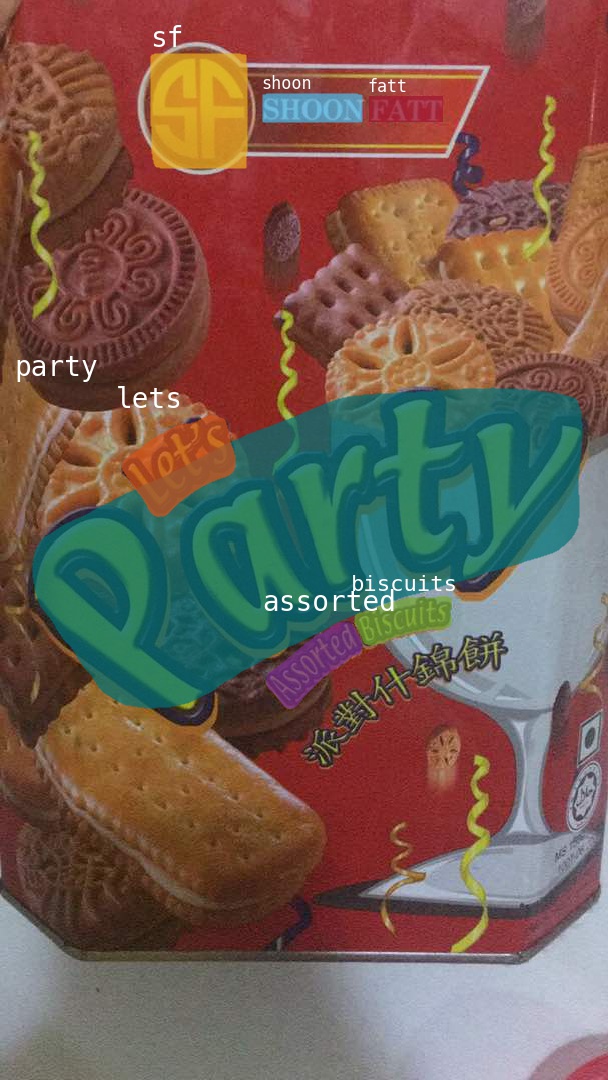}\label{fig:swint}}
    \caption{Qualitative analysis of SwinTextSpotter and other existing methods. Best view in screen.}
    \label{fig:qualitative_comparisons_results}
\end{figure}

We make some qualitative analysis with previous method which is shown in Fig \ref{fig:qualitative_comparisons_results}. It can be seen that previous methods failed on the difficult text instance such as ``Party'', while SwinTextSpotter can handle such case by exploiting the synergy of text detection and recognition. 

Intuitively, 
the detection result of SwinTextSpotter is more accurate.

\begin{table}[h]
\begin{tabular}{c|cc}
\hline
\multirow{2}*{Method}   & \multicolumn{2}{c}{Total-Text}            \\ \cline{2-3} 
                          & \multicolumn{1}{c|}{Det-Hmean} & E2E-Hmean \\ \hline
SwinTextSpotter-withou RC & \multicolumn{1}{c|}{82.8}      & 63.4      \\
SwinTextSpotter           & \multicolumn{1}{c|}{83.2}      & 66.9      \\ \hline
\end{tabular}
\caption{Ablation study on Recognition Conversion.}
\label{ABRC}
\end{table}

\section{Ablation study of Recognition Conversion}
We verify the effectiveness of other components without RC which helps to better reveal the effectiveness of RC.

From Table \ref{ABRC}, the performance, that using other components without RC, drops from 83.2$\%$ to 82.8$\%$ for detection and 66.9$\%$ to 63.4$\%$ for end-to-end scene text spotting.

\section{Comparison different backbone in different frameworks}
We also try to replace ResNet50 with Swin-Transformer on ABCNet. From Table \ref{ComBK}, the result can be improved by 1.8$\%$ with Swin-Transformer in text spotting. But there is no improvement for detection. It is similar to the case in SwinTextSpotter.

\begin{table}[h]
\begin{center}
\caption{Comparison different backbone on different architectures. ABC-R50 means ABCNet with ResNet50. ABC-Swin means ABCNet with SwinTransformer. Det. means detection result. E2E means end-to-end text spotting result.}
\begin{tabular}{cc|cc|cc|cc}
\hline
\multicolumn{2}{c|}{ABC-R50}     & \multicolumn{2}{c|}{ABC-Swin}    & \multicolumn{2}{c|}{Our-R50}     & \multicolumn{2}{c}{Our-Swin}     \\ \hline
\multicolumn{1}{c|}{Det.} & E2E  & \multicolumn{1}{c|}{Det.} & E2E  & \multicolumn{1}{c|}{Det.} & E2E  & \multicolumn{1}{c|}{Det.} & E2E  \\ \hline
\multicolumn{1}{c|}{86.0} & 67.1 & \multicolumn{1}{c|}{86.0} & 68.9 & \multicolumn{1}{c|}{87.2} & 72.4 & \multicolumn{1}{c|}{87.3} & 74.0 \\ \hline
\end{tabular}
\label{ComBK}
\end{center}
\end{table}

{\small
\bibliographystyle{ieee_fullname}
\bibliography{egbib.bib}
}
\end{document}